\documentclass[11pt]{article}
\usepackage[T1]{fontenc}
\usepackage[utf8]{inputenc}
\usepackage{acl}
\usepackage{times}
\usepackage{kotex}
\usepackage{amssymb}
\usepackage{pifont}

\usepackage{hyperref}
\usepackage{url}
\usepackage{graphicx} 
\usepackage{booktabs}
\usepackage{amsmath}
\usepackage{multirow}
\usepackage{algpseudocode}
\usepackage{algorithm}
\usepackage{tikz}
\usepackage{subcaption,comment,color,soul}
\usepackage[most]{tcolorbox}
\tcbuselibrary{minted}
\usepackage{listings}
\usepackage{array}
\usepackage{stfloats}
\usepackage{graphicx}
\usepackage{float}
\usepackage{wrapfig}

\usepackage[dvipsnames,table]{xcolor}
\usepackage[colorinlistoftodos,prependcaption,obeyFinal]{todonotes}

\definecolor{ruby}{RGB}{242,0,20}
\definecolor{emerald}{RGB}{26,121,42}
\definecolor{topaz}{RGB}{236,185,57}
\definecolor{sapphire}{RGB}{14,26,164}
\definecolor{flax}{rgb}{0.93, 0.86, 0.51}

\presetkeys{todonotes}{inline, color=topaz!40, bordercolor=topaz}{}

\newtcblisting{pythoncode}{
  listing engine=minted,
  listing only,
  nobeforeafter,
  minted style=colorful,
  minted language=python,
  minted options = {
        breaklines=true, 
        fontsize=\small
      },
}

\newtcblisting{javacode}{
  listing engine=minted,
  breakable,
  listing only,
  minted style=colorful,
  minted language=java,
  enhanced,
  minted options = {
        breaklines=true, 
        fontsize=\small
      },
}

\title{ContractEval: A Benchmark for Evaluating Contract-Satisfying Assertions in Code Generation}

\author{
Soohan Lim\textsuperscript{1,*},
Joonghyuk Hahn\textsuperscript{1,*},
Hyunwoo Park\textsuperscript{2},
Sang-Ki Ko\textsuperscript{2},
Yo-Sub Han\textsuperscript{1,$\dagger$}
 \\
\textsuperscript{1}Yonsei University, Seoul, Republic of Korea
\\
   \texttt{\{%
   \href{mailto:aness1219@yonsei.ac.kr}{aness1219},%
   \href{mailto:greghahn@yonsei.ac.kr}{greghahn},%
   \href{mailto:emmous@yonsei.ac.kr}{emmous}%
   \}@yonsei.ac.kr}
\\
\textsuperscript{2}University of Seoul, Seoul, Republic of Korea \\
   \texttt{\{%
   \href{mailto:hwpark03@uos.ac.kr}{hwpark03},%
   \href{mailto:sangkiko@uos.ac.kr}{sangkiko}%
   \}@uos.ac.kr}
}

\begin{document}

\maketitle

\begingroup
\renewcommand\thefootnote{\fnsymbol{footnote}}
\footnotetext[1]{Equal contribution.}
\footnotetext[2]{Corresponding author.}
\endgroup

\begin{abstract}
Current code generation evaluation measures functional correctness on well-formed inputs that satisfy all input preconditions.
This paradigm has a critical limitation: 
task descriptions often leave these preconditions implicit, while evaluation filters out inputs that violate them.
As a result, generated code may achieve high pass@k scores while failing to enforce the preconditions that the task actually requires. 
To address this gap, we introduce \textbf{ContractEval}, 
a benchmark for evaluating whether generated code enforces such preconditions---commonly referred to as contracts.
Built on HumanEval+ and MBPP+, 
ContractEval consists of 364 tasks, each with three components:
(i) descriptions reconstructed to explicitly state the contracts,
(ii) test cases synthesized through a neuro-symbolic pipeline that pairs an LLM with an SMT solver
to evaluate whether generated code satisfies these contracts, and
(iii) reference code combined with contracts.
Using ContractEval to evaluate five representative open-source code LLMs,
we reveal a stark disparity between functional correctness and contract satisfaction.
Under standard prompting, these models achieve
pass@1 of 75--82\% with 0\% contract satisfaction.
Even when contracts are explicitly stated in the prompt, the satisfaction rate reaches only 23--41\%.
This indicates that current LLMs struggle to satisfy contracts in their generated code,
establishing contract satisfaction as a crucial and previously overlooked axis of code generation quality.
Our code is available at \url{https://github.com/suhanmen/ContractEval}.
\end{abstract}

\section{Introduction}\label{sec:intro}
Current code generation benchmarks~\citep{Chenetal21,humaneval_and_mbpp,
LiuXW023, Shuyin} have driven strong progress on functional correctness.
They measure pass@k~\citep{KulalPC0PAL19,Chenetal21} over well-formed inputs.
Test cases satisfy all input preconditions by construction.
Evaluation then judges generated code on input--output correspondence alone.

This paradigm leaves a critical blind spot. Real-world software
must reject \emph{ill-formed inputs}: invalid data that violate
input validity constraints. Otherwise, invalid inputs slip through
and cause failures deep inside the program. In practice, developers
widely enforce these constraints with \texttt{assert} statements at
function entry. 
A statistical analysis of the most starred Python repositories on
GitHub~(averaging 60,898 stars) reveals that 22\% include input-validation
assertions inside their functions, indicating that rejecting
ill-formed inputs via assertions is a standard defensive practice.
Following Design by Contract~\citep{Meyer92}, we call such
assertion-level checks \emph{contracts}. 
Despite their prevalence in real-world code, 
contracts are largely overlooked by current benchmarks. 
Surprisingly, this issue persists even in benchmarks 
that acknowledge such contracts.
HumanEval+ and MBPP+~\citep{LiuXW023} already annotate each task with contracts as
a dedicated dataset field. Yet their evaluation suites filter out
every contract-violating input by construction. 
Contract satisfaction thus remains unmeasured.

We introduce \textbf{ContractEval}, a benchmark that fills this gap.
ContractEval evaluates whether the generated code implements contract-satisfying
assertions that reject ill-formed inputs. 
It extends HumanEval+ and MBPP+ by augmenting each task with three additions: 
(i) a contract-aware description stating input constraints from reference assertions, 
(ii) a set of contract-violating tests~(CVTs) that violate a specific contract clause
while remaining feasible under the rest, 
and (iii) reference code combined with these contracts.
We construct CVTs through a neuro-symbolic pipeline.
An LLM translates assertions into SMT constraints.
An SMT solver~\citep{smt} then enumerates satisfiable violation
combinations, yielding precisely targeted test inputs.

Using ContractEval, five code LLMs
achieve 0\% contract satisfaction with standard prompting.
Their pass@1 on the same tasks ranges from 75\% to 82\%.
Programs that look correct silently accept inputs any robust implementation
should reject. This gap is an illusion of correctness that pass@k cannot detect.
Contract satisfaction also shifts under prompt-level interventions.
Even when contracts are explicitly stated in the prompt, the satisfaction rate reaches only 23--41\%.
However, showing models a small set of CVT examples lifts contract satisfaction further to 49--53\%.
This improvement preserves 92\% of the original pass@1.
These findings indicate that current LLMs still struggle to satisfy contracts in their generated code. 
By exposing this blind spot, ContractEval establishes contract satisfaction as a crucial, 
yet previously overlooked, axis of code generation quality.


\section{Related Work}\label{sec:related}


\subsection{Code Generation Evaluation}\label{ssec:related-test-and-code}
Most code generation benchmarks evaluate models by executing
generated programs on unit tests and reporting
pass@k~\citep{testeval_naacl,Testgeneval_ICLR}, which captures
functional correctness on
well-formed inputs~\citep{Chenetal21,HendrycksBKMAGB21,Lietal22}.
Recent extensions~\citep{ChenZNZLLC23,LiuXW023} improve coverage by
adding stronger or more diverse valid tests, yet their evaluation
suites still target only well-formed inputs.
In particular, HumanEval+ and MBPP+~\citep{LiuXW023} annotate each
task with a dedicated contract field, but use it solely to
\emph{filter out} contract-violating inputs from the test suite
rather than to evaluate rejection behavior. As a result, even
benchmarks that already encode validity constraints leave the
rejection axis entirely unmeasured, and whether a generated program
rejects ill-formed inputs that violate those constraints remains an open question. 
While adjacent domains like hardware design have recently 
begun evaluating LLM-generated assertions~\citep{PulavarthiNDP25}, 
this type of assertion-aware evaluation remains absent in software code generation.
ContractEval addresses this critical blind spot in software benchmarks 
by introducing a contract-satisfaction axis that executes 
contract-violating inputs rather than filtering them out from the evaluation process.

\subsection{Contract Validity and Robustness}\label{ssec:related-contract}
Design by Contract (DbC) treats preconditions, postconditions, and invariants as part of the specification,
and emphasizes \emph{checking} them in executable conditions such as assertions~\citep{Meyer92}.
Contracts most commonly appear as assertion-level input validity checks that delimit the intended input space.
Recently, this DbC paradigm has been integrated into LLM-based program synthesis,
enabling automated contract inference~\citep{Greiner}, natural language-based formal contract synthesis~\citep{ye2026intentaligned} and test assertion generation~\citep{PrimbsFF25},
while utilizing explicit contracts as design constraints to improve accuracy~\citep{NewcombNO25}.
Despite this growing interest, prior work on robustness-oriented testing studies \emph{failure-inducing} inputs
that trigger generic exceptions~\citep{Exlong, Zhong_2025}.
However, a crashing input does not necessarily correspond to any particular contract assertion.
For instance, consider a function that requires a list of positive numbers.
An input \verb|None| may trigger a \verb|TypeError| regardless of the intended rule, ``all numbers must be positive.''
In contrast, contract evaluation requires tests that intentionally violate a chosen contract while remaining feasible with respect to others.
This distinction highlights the need for systematic methods that can precisely target formal contract specifications rather than just triggering arbitrary errors.
ContractEval targets this assertion-level notion of contracts and
evaluates whether the generated programs reject ill-formed inputs as intended,
using reference assertions in HumanEval+ and MBPP+ as a starting point.

\begin{figure*}[t]
    \centering
     \includegraphics[
        width=.97\textwidth,
        trim=0 0 50 0,
        clip
        ]
        {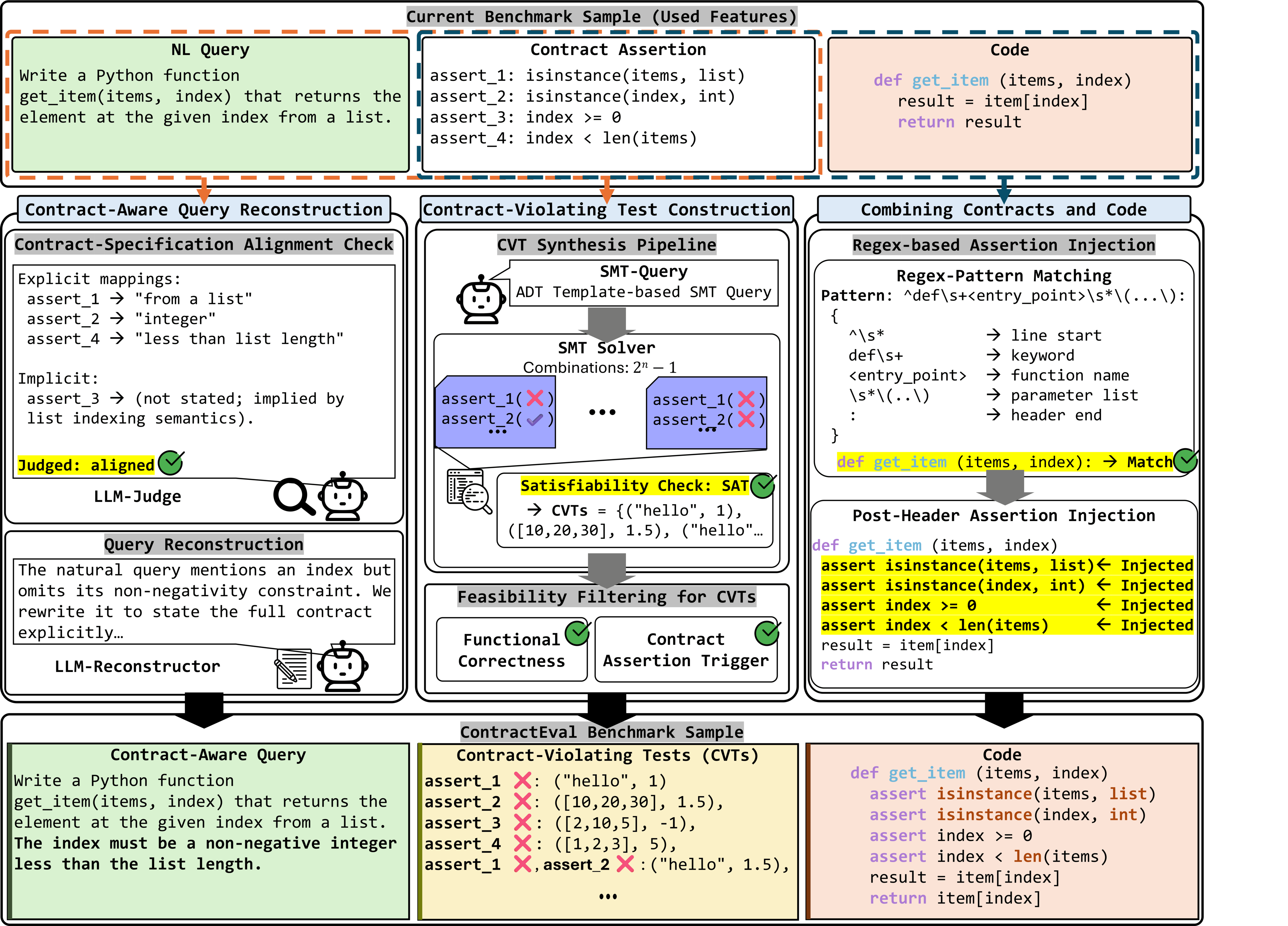}
    \caption{
    An overview of our ContractEval construction.
    }
    \label{fig:contracteval-construction}
\end{figure*}

\section{ContractEval Benchmark}\label{sec:method}
ContractEval extends each HumanEval+ and MBPP+~\citep{LiuXW023} task
with three additions: 
(i)~a contract-aware query that states input constraints explicitly, 
(ii)~a set of verified CVTs derived from reference assertions, and 
(iii)~reference code combined with these contracts.
Figure~\ref{fig:contracteval-construction} illustrates the
overall construction pipeline. We also define metrics that quantify
both CVT quality and contract satisfaction of generated code.

\subsection{Contract-Aware Query Reconstruction}\label{ssec:method-query-alignment}

\paragraph{Contract--Specification Alignment Check.}
Contract satisfaction evaluation assumes that each contract captures
input constraints either stated or implied by the task description.
However, irrelevant, overly restrictive, or contradictory contracts make
evaluation noisy rather than informative.
We therefore assess contract--specification alignment before any downstream analysis
using an LLM judge that reads each task's description together with
its reference contracts and decides whether they are consistent.
Across 542 tasks that include reference assertions, 88.04\%
are judged specification-aligned. Tasks with misaligned
contracts are excluded from the benchmark.

\paragraph{Query Reconstruction.}
For each retained task, we reconstruct a contract-aware query that
restates the reference assertions as explicit input constraints. 
A commercial LLM interprets each assertion and
rewrites the original query to include the corresponding constraint.
This step standardizes how contract information is presented to code
generation models and enables controlled comparisons that isolate a
model's ability to satisfy explicitly stated constraints.

\subsection{Contract-Violating Test Construction}\label{ssec:method-cvt}
\paragraph{CVT Synthesis Pipeline.}
We propose a two-stage neuro-symbolic pipeline that combines the 
semantic translation capability of LLMs with the mathematical guarantees 
of SMT solvers for strictly satisfying logical constraints.
The pipeline produces CVTs that violate a chosen subset of contracts while
remaining feasible under the rest.
The first stage converts contract assertions
into SMT-LIB constraints over an Algebraic Data Type~(ADT).
The second stage enumerates contract subsets,
checks their satisfiability, and synthesizes concrete inputs 
from the resulting combinations of constraints, yielding targeted CVTs.

\paragraph{Formalization with a Canonical ADT Template.}
Programs may accept inputs that combine primitive and structured values.
These include nested containers with mixed element types.
Such inputs are hard to encode in SMT directly.
Most solvers assume statically typed signatures and
require explicit constructors for structured data.
We address this challenge with a canonical value ADT embedded in an SMT-LIB template,
as detailed in Appendices~\ref{app:smt-template}--\ref{app:smt-template-case-study}.
The ADT provides a unified representation for integers, floats, strings, booleans,
and nested containers.
Contracts are then encoded as SMT constraints
over this single representation using shared constructors and auxiliary predicates.
This uniform formalization supports the systematic construction of constraint sets
for any targeted violation combination across all contract subsets.
An LLM automates this process by translating reference contracts into ADT-level constraints. 
The resulting SMT-LIB instance is syntactically strict and preserves the semantics required by the solver.

\paragraph{Combinatorial Synthesis via SMT Solving.}
The pipeline generates test cases that cover distinct and logically feasible violation scenarios.
Specifically, for a task with $n$ contracts, we consider all $2^n - 1$ non-empty subsets as potential violation targets.
For each target subset, we construct an SMT query that deliberately negates the targeted 
contract predicates while enforcing the satisfaction of all remaining predicates.
The solver then evaluates these queries, discarding \texttt{UNSAT} targets that 
logically conflict with the enforced constraints.
For each \texttt{SAT} query, the solver returns an ADT-level model 
assignment, which constitutes a CVT within the encoded space.
Finally, we decode this assignment into an executable test input by
converting ADT constructors into runtime values and recursively rebuilding nested containers.
This process yields CVTs that are both precisely targeted and logically consistent, 
which then proceed to the feasibility filtering stage.

\paragraph{Feasibility Filtering for CVTs.}
Test cases synthesized by the SMT solver are logically consistent
with the contract predicates. Logical consistency alone does not
ensure that an input is suitable for evaluating contract
satisfaction. A valid CVT must be functionally feasible while
violating the intended contract. The same input should run
successfully when contract checks are removed, and should then fail
with the intended assertion when those checks are enabled. 
If an input instead crashes with a generic runtime exception, such as \texttt{TypeError}
or \texttt{IndexError}, it reflects functional invalidity rather than
contract enforcement.

Accordingly, we filter synthesized CVTs by executing each input on the reference
implementation under two configurations: (i)~with assertions
removed to verify functional feasibility, and (ii)~with assertions
intact to confirm the targeted violation. 
As a consequence of this input-level filtering, tasks that fail to retain 
any functionally valid CVTs are also excluded from the benchmark.
Combined with the alignment filter from
Section~\ref{ssec:method-query-alignment}, 364 tasks remain for
downstream evaluation. The retained tasks ensure that ContractEval
measures contract enforcement rather than robustness to errors.

\subsection{Combining Contracts and Code}\label{ssec:contracteval-code}
We integrate the reference assertions into each task's canonical implementation 
to create a contract-aware reference program. 
This program strictly enforces every validity constraint at the function entry. 
Specifically, we locate the target function header using a regular expression 
anchored to the specified function name.
This robust matching reliably identifies the correct signature, even when helper definitions,
imports, or decorators appear in the same file.
Immediately after the matched header, we insert the task's assertion-level contracts
as a contiguous block preceding the original function body.
The resulting combined program preserves the functional 
behavior of the reference implementation on well-formed inputs and raises an 
\texttt{AssertionError} on any input that violates a contract clause.

\subsection{ContractEval Metrics}\label{ssec:contracteval-metrics}
ContractEval reports four metrics.
Assertion violation capture~(AVC) and target specificity~(TS) validate the CVTs themselves.
pass@k measures functional correctness on well-formed inputs.
Contract satisfaction rate~(CSR) measures whether generated code rejects CVTs.

\paragraph{CVT Quality Metrics~(AVC, TS).}
Let $A=\{a_1,\ldots,a_n\}$ be the set of contract assertions and $T=\{t_1,\ldots,t_m\}$ be the generated test cases.
For a test case $t\in T$, let $F_t\subseteq A$ denote the set of assertions 
violated when executing $t$.
AVC measures the fraction of generated test cases that successfully trigger at least one assertion violation:
\[
\mathrm{AVC}
=
\frac{|\{t \mid t \in T \wedge |F_t|>0\}|}{|T|}.
\]
A higher AVC indicates that the generator more consistently produces CVTs that 
elicit at least one assertion failure.

TS evaluates how each CVT matches its intended contract targets.
Let $T_{\mathrm{neg}}=\{t \in T \mid |F_t|>0\}$ be the set of tests that trigger at least one assertion violation.
For each $t \in T_{\mathrm{neg}}$, let $V_t \subseteq A$ denote the set of assertions that $t$ is intended to violate.
TS measures the Jaccard similarity between the intended target set $V_t$ and the actually violated set $F_t$ and averages it over $T_{\mathrm{neg}}$:
\[
\mathrm{TS}
=
\frac{1}{|T_{\mathrm{neg}}|}
\sum_{t \in T_{\mathrm{neg}}}
\frac{|F_t \cap V_t|}{|F_t \cup V_t|}.
\]
A TS of $1$ indicates that each test violates exactly its intended target set.
A smaller TS indicates that a test misses some intended violations, triggers unintended violations, or both.

\paragraph{Contract Satisfaction Rate~(CSR)}
To evaluate contract enforcement against contract-violating inputs,
we use verified CVTs obtained from our filtering process.
Let $T_{\mathrm{ver}}=\{t_1,\ldots,t_k\}$ denote the verified CVT set.
For a generated code snippet $C$,
let $T_{\mathrm{rej}}\subseteq T_{\mathrm{ver}}$ be the subset of inputs that $C$
rejects by raising an assertion failure or an explicit rejection.
CSR quantifies the fraction of verified CVTs that are rejected by $C$:
\[
\mathrm{CSR}
=
\frac{|\{t \mid t \in T_{\mathrm{ver}} \wedge t \in T_{\mathrm{rej}}\}|}{|T_{\mathrm{ver}}|}.
\]
A higher CSR indicates that the generated code more reliably enforces the specified 
contracts by successfully rejecting contract-violating inputs.

\section{Experimental Settings}\label{sec:exp}


\subsection{Datasets}\label{ssec:dataset}
We evaluate on ContractEval built on HumanEval+ and MBPP+~\citep{LiuXW023}. 
For each task, we use the reference
well-formed tests to measure functional correctness and a set of
verified CVTs to measure contract satisfaction. 
Table~\ref{tab:dataset_statistics} summarizes the benchmark
statistics: ContractEval comprises 364 tasks with an average of 2.29
assertions per task, 2.89 satisfiable violation combinations per
task, and a total of 4,907 verified CVTs after feasibility filtering.


\subsection{Prompting Conditions}\label{ssec:prompting_cod}

We probe ContractEval under three prompting conditions to assess
how sensitive contract satisfaction is to the information available
in the prompt. The \emph{standard} condition uses the reference
natural-language query. \emph{Contract specification}~(CS) uses the
contract-aware rewrite from
Section~\ref{ssec:method-query-alignment}. \emph{Example-augmented
specification}~(EAS) further appends verified CVTs
as concrete examples of inputs that should be rejected.

\subsection{Evaluation Metrics}\label{ssec:eval_metric}

ContractEval introduces three dedicated metrics as detailed in
Section~\ref{ssec:contracteval-metrics}. AVC and TS validate the
quality of generated CVTs. CSR directly measures whether generated
code rejects contract-violating inputs by executing CVTs against the
implementation, providing a definitive signal of contract
satisfaction. We retain \texttt{pass@k} on well-formed tests to track
functional correctness. As supplementary analysis, we additionally
report \textsc{CodeBLEU}~\citep{abs-2009-10297} and an
\emph{LLM-as-judge} score, which approximate the syntactic and
semantic quality of generated assertion checks but do not verify
actual rejection behavior.


\subsection{Evaluated Models}\label{ssec:evaluated_models}
We employ \texttt{o4-mini} as the primary engine for contract-violating test case generation and
contract-aware query reconstruction, and as the LLM-as-judge.
We justify selecting \texttt{o4-mini} over larger models in Appendix~\ref{app:llm_engine_selection}.
For our code generation experiments, we evaluate a set of open-source models
including
gemma-3-12B-it~(gemma-3),
DeepSeek-R1-Distill-Qwen-14B~(DeepSeek-R1),
Qwen3-14B~(Qwen-3),
Phi-4-reasoning~(Phi-4), and
Phi-4-reasoning-plus~(Phi-4-plus).
We provide detailed implementation settings in Appendix~\ref{ap:implementation-detail} for completeness and reproducibility.

\begin{table}[t]
\caption{Distributional statistics of contracts and generated CVTs
in the ContractEval benchmark.}
\label{tab:dataset_statistics}
\centering
\begin{tabular}{p{0.7\columnwidth} r}
\toprule
\textbf{Metric} & \textbf{Value} \\
\midrule
Average Assertions per Task & 2.29 \\
Average SAT Subsets per Task & 2.89 \\
Average CVTs per Task & 13.48 \\
Total Verified CVTs (Filtered) & 4,907 \\
\bottomrule
\end{tabular}
\end{table}

\section{Main Results}\label{sec:result}
We use ContractEval to measure how well current LLMs generate
contract-checking assertions alongside the functional body.
Functional correctness is reported as pass@1 on well-formed tests,
and contract satisfaction is reported as CSR on verified CVTs.
We compare three prompting conditions from
Section~\ref{ssec:prompting_cod}: Standard, CS, and EAS.

\begin{table*}[t]
\caption{Functional correctness and contract satisfaction in code generation.
Adding contract descriptions~(CS) helps, and adding a few ill-formed examples~(EAS) yields large gains in contract satisfaction.}
\label{tab:full_result}
\centering
\begin{tabular}{llrrrr}
\toprule
\multirow{2}{*}{\textbf{Model}} & \multirow{2}{*}{\textbf{Instruction}} & \multicolumn{1}{c}{\textbf{Functional}} & \multicolumn{3}{c}{\textbf{Contract Satisfaction}} \\
\cmidrule(lr){3-3} \cmidrule(lr){4-6}
 & & \multicolumn{1}{c}{pass@1 ($\uparrow$)} & \multicolumn{1}{c}{CSR ($\uparrow$)} & \multicolumn{1}{c}{CodeBLEU ($\uparrow$)} & \multicolumn{1}{c}{LLM-as-judge ($\uparrow$)} \\\midrule
\multirow{3}{*}{DeepSeek-R1}    & Standard         & \textbf{80.81}\%        & 0.00\%                  & 0.00\%                       & 0.00\%                \\
                                & CS               & 73.29\%                 & 27.24\%                 & 34.29\%                      & 53.05\%               \\
                                & EAS              & 70.87\%                 & \textbf{52.66}\%        & \textbf{54.51}\%             & \textbf{81.49}\%      \\\midrule
\multirow{3}{*}{gemma-3}        & Standard         & \textbf{81.76}\%        & 0.00\%                  & 0.00\%                       & 0.00\%                \\
                                & CS               & 74.75\%                 & 22.69\%                 & 29.97\%                      & 41.74\%               \\
                                & EAS              & 69.56\%                 & \textbf{48.54}\%        & \textbf{48.88}\%             & \textbf{72.04}\%      \\\midrule
\multirow{3}{*}{Phi-4}          & Standard         & \textbf{75.39}\%        & 0.00\%                  & 0.00\%                       & 0.00\%                \\
                                & CS               & 71.29\%                 & 40.71\%                 & 46.66\%                      & 69.33\%               \\
                                & EAS              & 72.99\%                 & \textbf{50.73}\%        & \textbf{50.86}\%             & \textbf{75.63}\%      \\\midrule
\multirow{3}{*}{Phi-4-plus}     & Standard         & \textbf{75.97}\%        & 0.00\%                  & 0.00\%                       & 0.00\%                \\
                                & CS               & 70.83\%                 & 38.24\%                 & 44.38\%                      & 65.31\%               \\
                                & EAS              & 73.21\%                 & \textbf{50.94}\%        & \textbf{51.04}\%             & \textbf{75.98}\%      \\\midrule
\multirow{3}{*}{Qwen-3}         & Standard         & \textbf{74.79}\%        & 0.00\%                  & 0.00\%                       & 0.00\%                \\
                                & CS               & 70.62\%                 & 24.29\%                 & 35.05\%                      & 49.71\%               \\
                                & EAS              & 69.65\%                 & \textbf{51.83}\%        & \textbf{53.30}\%             & \textbf{77.27}\%      \\
\bottomrule
\end{tabular}
\end{table*}

\paragraph{Standard Prompting Produces an Illusion of Correctness.}
Table~\ref{tab:full_result} shows that the standard prompt yields 0\% CSR
for every base LLM despite 75--82\% pass@1.
The models implement the requested functionality, yet they do not reject
even a single CVT.
We read this as evidence that standard
natural-language task descriptions lead models
toward implementing functionality without input validation.
Writing a function that ``returns the number of odd digits in a list of strings''
invites an implementation rather than a rejection policy.
The description says nothing about what types or ranges the inputs must
satisfy, so the model has no textual cue to insert an input-validation check.
Because pass@k only executes well-formed inputs, the absence of input validation
slips past the evaluation.
ContractEval exposes this gap by executing CVTs that a
contract-respecting implementation should reject rather than silently accept.

\paragraph{Explicit Contracts Close Most of the Gap.}
CS lifts CSR by 23--41\%p over the standard prompt.
The range is wide and uneven across models.
The variation suggests that translating the contract clauses
surfaced in CS into
assertion code depends heavily on how the model parses modifiers such as
``positive integer'' or ``non-empty list''.
EAS yields the largest gains, adding another
10--30\%p on top of CS. On average, EAS achieves 50.94\%p CSR improvement
over the standard prompt and preserves 92\% of the original pass@1.
The preserved pass@1 deserves attention.
It indicates that contract checks are largely additive to the functional
body rather than a substitute.
The small pass@1 drop therefore reflects a mild interference effect,
not a structural trade-off between them.

\paragraph{Auxiliary Metrics Agree with CSR.}
CodeBLEU and LLM-as-judge both score 0\% under the standard prompt,
because the generated code contains no contract-checking assertions to compare against.
Under EAS, the two scores exceed CS by 13.65\%p and 20.65\%p.
The LLM-as-judge gap is the larger of the two
because LLM-as-judge evaluates whether generated assertions check
the same conditions as the reference,
while CodeBLEU measures token overlap.
In Python, assertion syntax is compact enough that semantically
different checks can share most of their tokens, which inflates
CodeBLEU and narrows its gap. The wider LLM-as-judge gap, therefore,
suggests that EAS assertions check the right conditions, not merely
that they are present.

\paragraph{Contract Satisfaction Does not Saturate, Even Under EAS.}
When we restrict the evaluation to tasks where the generated program
already scores pass@1$=100\%$, the mean CSR lies in the 51\%--59\% range across models.
Even when a program solves the task correctly, it still misses close
to half of the contract clauses that the reference enforces. 
A perfect pass@1 does not entail a high CSR 
because the two metrics capture different properties: pass@1 tests whether the function computes the
right output on valid inputs, while CSR tests whether it rejects
inputs that violate the contract. Solving the task and guarding its
boundary are separate objectives, and evaluating only the former
leaves the latter unmeasured.
We provide an analysis of this conditional behavior across models 
and prompting conditions in Appendix~\ref{app:contract-evaluation-correct-function}.

\begin{table}[t]
\caption{Alignment rate of contract specification judgments across
different evaluators on a 100-task sample.}
\label{tab:alignment_rate}
\centering
\begin{tabular}{lr}
\toprule
\textbf{Evaluator} & \textbf{Alignment Rate} \\
\midrule
Human evaluation & 92.00\% \\
GPT-4o & 96.48\% \\
Gemini-3-flash & 97.80\% \\
\bottomrule
\end{tabular}
\end{table}

\section{Analysis}
Our analysis has two goals.
Sections~\ref{ssec:analysis-alignment} and~\ref{ssec:analysis-cvt}
examine whether ContractEval itself is trustworthy as a
measuring instrument for contract satisfaction,
covering the reliability of the alignment filter and the precision of the generated CVTs.
Sections~\ref{ssec:analysis-ablation},~\ref{ssec:rq-contract-ct-in-codegen}, and~\ref{ssec:analysis-commercial}
then use the benchmark to interpret how code generation models actually behave.
Section~\ref{ssec:analysis-ablation} shows that functional failure
demonstrations alone do not induce contract checks,
no matter how many are added to the prompt.
Section~\ref{ssec:rq-contract-ct-in-codegen} then asks whether
popular prompting techniques layered on top of EAS improve the
pass@1 versus CSR trade-off.
Section~\ref{ssec:analysis-commercial} extends the study to
commercial LLMs and shows that contract failure persists at scale.

\subsection{Alignment Reliability}\label{ssec:analysis-alignment}
Contract satisfaction evaluation only makes sense when the
reference assertions capture what the task description
actually asks for.
HumanEval+ and MBPP+ provide reference assertions as a dataset field, but
they do not guarantee semantic consistency between those assertions and
the natural language task description.
We filter tasks using an LLM judge in
Section~\ref{ssec:method-query-alignment}.
A single judge model may be biased,
so we validate this filter with two independent checks:
human evaluation and cross-model agreement.

Two computer science experts assessed the semantic
alignment on a 100-task sample
and resolved disagreements by discussion.
They agreed with the filter on 92.00\% of tasks. 
The remaining 8.00\% were cases where the original assertions
over-constrained vague natural-language prompts.
For example, while the text might broadly ask for ``a list'',
the assertion explicitly narrows the element types down to integers.
Re-running the filter with GPT-4o and Gemini-3-flash produced alignment rates
of 96.48\% and 97.80\%, as summarized in
Table~\ref{tab:alignment_rate}.
The three rates---from one human pair and two LLM judges---fall within a 6\%p band.
The band is narrow enough to treat
the filtered task set as reflecting a shared judgment across evaluators
rather than an artifact of any single judge.

\begin{table}[t]
\caption{Comparison of contract-violating test cases generated
by \texttt{o4-mini} and by ContractEval~(ours).}
\label{tab:testcase_generation_comparison}
\centering
\resizebox{\linewidth}{!}{
\setlength{\tabcolsep}{3pt}
\begin{tabular}{lcccc}
\toprule
\textbf{Method} 
& \textbf{AVC}~(${\uparrow}$) 
& \textbf{TS}~(${\uparrow}$)
& \textbf{Average}~(${\uparrow}$) 
\\ \midrule
    o4-mini               & \textbf{97.58}\%  & 72.69\%           & 85.13\%          \\
    ContractEval~(ours)   & 94.11\%           & \textbf{84.92}\%  & \textbf{89.52}\% \\ 
    \bottomrule
\end{tabular}
}
\end{table}

\subsection{Precision of Generated CVTs}\label{ssec:analysis-cvt}
A useful CVT violates
exactly the contracts it is supposed to violate.
If it trips additional unrelated assertions,
or fails before any assertion fires,
it measures something other than contract enforcement.
We compare two generators along this precision axis.
A direct \texttt{o4-mini} baseline generates
ill-formed tests without SMT assistance.
Our neuro-symbolic pipeline pairs an LLM
translation step with an SMT solver.

Table~\ref{tab:testcase_generation_comparison} summarizes the outcome.
Both approaches reach high AVC, with 97.58\% for the LLM baseline
and 94.11\% for ContractEval.
The more informative number is TS,
which rises from 72.69\% to 84.92\% under our pipeline.
The 12.23\%p gap indicates that our CVTs land on the
intended target set much more often than the baseline does.
When the LLM baseline reaches high AVC, it typically does so
by triggering whichever assertion is easiest to violate,
even if that assertion is not the chosen target.
The SMT step removes this ambiguity, because the solver
returns only inputs that satisfy the exact combination of violated
and satisfied clauses that we asked for.

The slight AVC drop for our pipeline is a direct consequence of
this precision. The LLM baseline spends most of its budget on easy-to-violate
clauses, which keeps AVC high but leaves harder target combinations untouched.
Our pipeline instead enumerates every feasible combination,
including those that require tightly constrained inputs
such as a non-empty list of positive integers with a bounded length.
Such combinations are intrinsically harder to satisfy, 
and some do not trigger any assertion.
This marginally lowers AVC but substantially broadens the space of
contract combinations that the benchmark actually covers.

We view this trade-off as favorable for contract evaluation.
Contract enforcement in real code must hold across interacting clauses,
not just the one that is easiest to violate.
A benchmark that pins down precise targets therefore gives a
cleaner signal of whether a generated program respects the full contract,
rather than whether it happens to catch the single weakest link.
A case study of baseline-specific failure patterns,
including logically contradictory target subsets produced by the
direct LLM approach, appears in Appendix~\ref{ap:qa1}.

\subsection{Functional Failure Demonstrations vs. Contract Checks}\label{ssec:analysis-ablation}
Contract-level prompting such as EAS lifts CSR from 0\% to roughly 50\%.
We now examine whether functional failure demonstrations---wrong input--output pairs
on well-formed inputs---can achieve a similar effect.

\begin{figure}[t]
    \centering
    \includegraphics[width=.98\columnwidth]{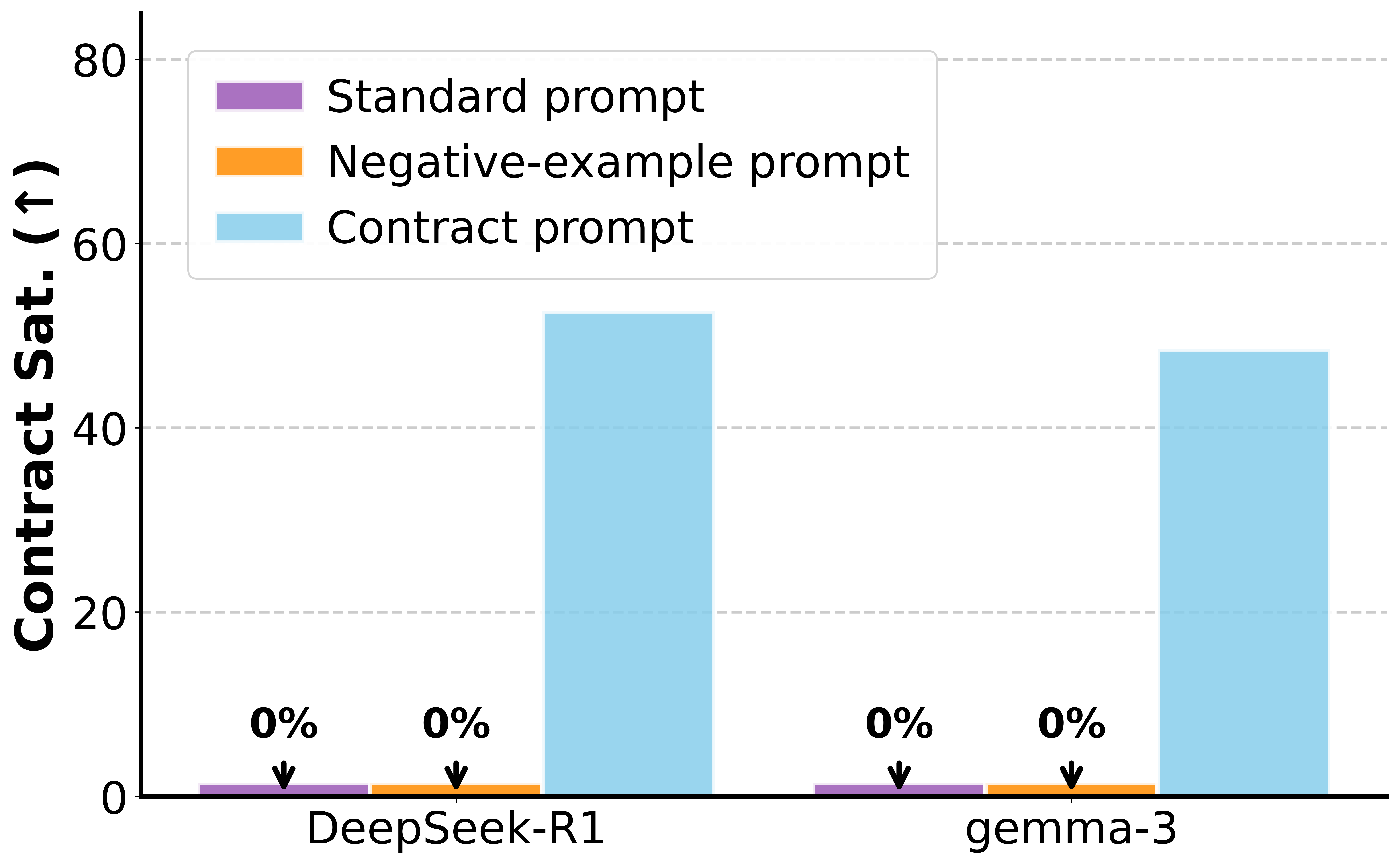}
    \caption{
    Contract satisfaction of generated code under three prompting
    settings. Standard and negative-example prompts yield 0\%, while
    the contract prompt achieves approximately 50\%.
    }
    \label{fig:obs_graph}
\end{figure}

We compare three prompting conditions.
The \emph{standard} prompt uses the reference natural-language
description.
The \emph{negative-example} prompt augments the standard prompt
with a small set of wrong input--output pairs for the same task.
Consider a function that should map the input~\texttt{0}
to the output~\texttt{1}.
A negative example takes the form ``input~\texttt{0},
output~\texttt{3}'', with the output explicitly marked as incorrect.
Such examples encode functional mistakes on well-formed inputs and
do not touch the input boundary.
The \emph{contract} prompt is our EAS setting:
a contract-aware description paired with CVTs.

Figure~\ref{fig:obs_graph} summarizes the outcome.
The negative-example prompt stays 0\% across models, indistinguishable
from standard, while the contract prompt lifts
CSR to roughly 50\%.
The reason is a mismatch of semantic layer.
A wrong-output example signals that a particular input--output mapping is incorrect.
It concerns the function's behavior on a \emph{valid} input.
The model may adjust the function body.
It has no reason to add an \texttt{assert} at the function entry,
because the input itself was never flagged as invalid.
Contract-level prompting operates at a different layer.
The contract-aware description names invariants in a form
that the model can translate into contract checks,
and the CVTs anchor those checks with concrete ill-formed inputs.
The two together carry boundary-level information, 
which a contract check actually encodes.
A bug report and a contract
violation look similar on the surface, but only the latter speaks
the contract language.


\begin{table}[t]
\caption{Comparison of additional prompting strategies.
Multi-turn recovers pass@1 while maintaining competitive contract satisfaction.
}
\label{tab:prompting_result}
\centering
\resizebox{.98\linewidth}{!}{
\setlength{\tabcolsep}{2.5pt}
\begin{tabular}{llcc}
\toprule
Model                                         & Instruction      & pass@1~(${\uparrow}$)        & CSR~(${\uparrow}$) \\\midrule
\multirow{3}{*}{DeepSeek-R1}                  & EAS              & 70.87\%                      & \textbf{52.66\%}   \\
                                              & + CoT            & 61.11\%                      & 35.00\%            \\
                                              & + Multi-Turn      & \textbf{72.96\%}             & 45.44\%            \\\midrule
\multirow{3}{*}{gemma-3}                      & EAS              & 69.56\%                      & 48.54\%            \\
                                              & + CoT            & 70.46\%                      & 50.37\%            \\
                                              & + Multi-Turn      & \textbf{71.34\%}             & \textbf{49.26\%}   \\\midrule
\multirow{3}{*}{Phi-4}                        & EAS              & \textbf{72.99\%}             & \textbf{50.73\%}   \\
                                              & + CoT            & 3.66\%                       & 1.91\%             \\
                                              & + Multi-Turn      & 71.88\%                      & 46.58\%            \\\midrule
\multirow{3}{*}{Phi-4-plus}                   & EAS              & 73.21\%                      & \textbf{50.94\%}   \\
                                              & + CoT            & 0.00\%                       & 0.00\%             \\
                                              & + Multi-Turn      & \textbf{77.42\%}             & 48.52\%            \\\midrule
\multirow{3}{*}{Qwen-3}                       & EAS              & 69.65\%                      & \textbf{51.83\%}   \\
                                              & + CoT            & 71.13\%                      & 45.24\%            \\
                                              & + Multi-Turn      & \textbf{71.70\%}             & 49.23\%            \\\bottomrule
\end{tabular}
}
\end{table}

\subsection{Effect of Prompting Strategies}\label{ssec:rq-contract-ct-in-codegen}
After establishing that EAS is effective for eliciting contract-aware behavior,
we analyze whether \emph{additional} prompting techniques improve the trade-off
between functional correctness and contract satisfaction.
We consider Chain-of-Thought~(CoT) prompting and a refinement-style multi-turn
strategy that first generates a solution with functional logic and then refines
the code with contract information. Table~\ref{tab:prompting_result} summarizes
the impact of these techniques across our base LLMs.

While CoT yields small pass@1 gains of under 1.5\%p for gemma-3 and Qwen-3,
it frequently degrades both pass@1 and CSR for the remaining models. 
More critically, CoT catastrophically fails on Phi-4 and Phi-4-plus, collapsing both pass@1 and
CSR to 0--4\%. This suggests that CoT is not appropriate in our setting, where
the output must remain executable and simultaneously reject CVTs as intended.

In contrast, the multi-turn strategy is stable across models and improves
functional correctness over EAS in most cases, raising pass@1 by 1.8\%p on
average. For instance, Phi-4-plus increases pass@1 from 73.21\% to 77.42\%,
and DeepSeek-R1 increases from 70.87\% to 72.96\%. This gain is typically
accompanied by a modest drop in CSR of 2--7\%p, indicating that refinement
tends to prioritize preserving functional behavior over implementing every
contract check.

Overall, EAS remains the strongest single-turn strategy for contract satisfaction, 
while the multi-turn strategy is the most reliable way to boost pass@1 on top of EAS.
None of the tested prompting variants achieves both high pass@1 and
near-perfect CSR, motivating future methods to be evaluated under ContractEval.

\subsection{Contract Failure in Commercial LLMs}\label{ssec:analysis-commercial}
Our main results use open-source models in the 12--14B range,
leaving open whether stronger commercial models behave differently
once contract information reaches the prompt. We evaluate GPT-4o
and Gemini-3-flash on a 100-task sample drawn from ContractEval
under the same three prompting conditions.

Table~\ref{tab:commercial_models_result} shows that commercial
models reproduce the open-source pattern rather than escaping it.
Under the standard prompt, GPT-4o and Gemini-3-flash reach 90.04\%
and 94.46\% pass@1 with 0\% CSR, confirming that without explicit
contract information in the prompt no model enforces contracts
regardless of scale. CS lifts CSR into the 36\% range for both
commercial models, and EAS extends this further to 56.51\% for
GPT-4o and 62.94\% for Gemini-3-flash. Even at larger scale with
contract-aware prompts, roughly 40\% of CVTs remain silently
accepted. Therefore, scale improves contract
satisfaction but does not eliminate the gap, and contract-aware
prompting remains necessary at every scale.

\begin{table}[t]
\caption{Functional correctness and contract satisfaction of
commercial LLMs on a 100-task sample.}
\label{tab:commercial_models_result}
\centering
\resizebox{\columnwidth}{!}{
\begin{tabular}{llrr}
\toprule
\textbf{Model} & \textbf{Instruction} & \textbf{pass@1} ($\uparrow$) & \textbf{CSR} ($\uparrow$) \\
\midrule
\multirow{3}{*}{GPT-4o} & Standard & \textbf{90.04\%} & 0.00\% \\
 & CS & 81.58\% & 36.09\% \\
 & EAS & 75.45\% & \textbf{56.51\%} \\
\midrule
\multirow{3}{*}{Gemini-3-flash} & Standard & \textbf{94.46\%} & 0.00\% \\
 & CS & 86.57\% & 36.02\% \\
 & EAS & 86.44\% & \textbf{62.94\%} \\
\bottomrule
\end{tabular}%
}
\end{table}

\section{Conclusion}\label{sec:conclusion}
ContractEval reframes code generation evaluation by testing not only functional correctness on well-formed inputs,
but also whether generated programs \emph{reject} CVTs via intended assertions.
Built on HumanEval+ and MBPP+, ContractEval augments each task with three components: 
(i) a contract-aware query that restates input constraints explicitly, 
(ii) a set of contract-targeted CVTs constructed through a neuro-symbolic pipeline that pairs an LLM with an SMT solver,
and (iii) a reference program that combines the canonical implementation with its assertion-level contracts.
ContractEval further introduces dedicated metrics to quantify both the precision of generated CVTs~(AVC and TS)
and the contract satisfaction of generated programs~(CSR). 
Empirically, we find that standard prompts for code generation yield 0\% contract satisfaction,
failing to reject CVTs.
This reveals a blind spot in conventional benchmarks.
By contrast, EAS raises mean contract satisfaction from 0\% to 50.94\% 
while preserving 92\% of the standard prompt's pass@1.
Overall, ContractEval provides a benchmark and methodology for evaluating assertion-level contract checks
as an essential axis of robustness in LLM code generation.

\section*{Limitations}\label{limitation}
ContractEval is designed to make assertion-level contract satisfaction measurable and comparable,
but several practical constraints define its current scope and point to natural extensions.

\paragraph{Scope of contracts.}
ContractEval derives contracts from reference assertions, which may be incomplete with respect to all validity constraints that developers intend.
We mitigate misalignment by filtering tasks whose assertions are judged inconsistent with the natural-language specification,
but ContractEval does not claim to exhaustively capture every invalid input boundary for each task.

\paragraph{Coverage of invalid inputs.}
CVTs are clause-targeted and feasibility-checked, but they still represent a finite set of contract-violation patterns.
They may not cover all semantically invalid inputs~(e.g., distributional shifts, adversarial edge cases, or domain-specific constraints not expressed as assertions).
Future versions could combine ContractEval’s targeted CVTs with complementary robustness tests (e.g., fuzzing-style perturbations) while preserving interpretability.


\paragraph{Cost and evaluation dependencies.}
SMT enumeration and feasibility filtering introduce computational overhead compared to purely LLM-based test generation,
and any auxiliary model-based judgments~(e.g., contract--specification alignment checks or semantic scoring) may inherit bias from the judge model.
Reducing reliance on model-based judgments and improving efficiency are practical directions for scaling ContractEval.
We detail the computational costs of our construction pipeline, 
including execution time measurements for the filtering process,
in Appendices~\ref{app:computational_cost}--\ref{app:feasibility_filtering}.

\section*{Acknowledgments}
This research was supported by the NRF grant~(RS-2025-00562134) and 
the AI Graduate School Program~(RS-2020-II201361) funded by the Korean government.

\bibliography{custom}

\appendix

\section{Computational Cost of CVT Generation}\label{app:computational_cost}
Table \ref{tab:time_cost} details the computational time cost for CVT generation across the entire dataset.
Generating the complete benchmark requires 126.89 minutes of o4-mini inference and 228.23 minutes of SMT solver operations.

While the SMT solver introduces additional construction time, this computational investment significantly enhances TS.
The SMT-based filtering pipeline achieves a TS score 12.23\%p higher than relying solely on o4-mini.
Rigorous feasibility checks guarantee the logical consistency of every synthesized test case.
Consequently, this substantial improvement in precise target specificity and absolute structural validity fully justifies the supplementary runtime cost.

\begin{table}[hbt]
\caption{Computational time cost for CVT generation across the entire dataset.}
\label{tab:time_cost}
\centering
\begin{tabular}{lr}
\toprule
\textbf{Model} & \textbf{Time Cost (min)} \\
\midrule
o4-mini & 126.89 \\
SMT solver & 228.23 \\
\bottomrule
\end{tabular}
\end{table}

\section{Efficiency of Feasibility Filtering}\label{app:feasibility_filtering}
Table \ref{tab:feasibility_filtering} reports the minimal computational time required for the final feasibility filtering phase.
The ContractEval pipeline completes this entire validation process in merely 1.75 minutes across the complete benchmark.
Such rapid execution demonstrates that the rigorous quality control step imposes virtually no overhead on the overall construction timeline.
Consequently, the pipeline seamlessly integrates strict feasibility checks without sacrificing operational efficiency.

\begin{table}[hbt]
\caption{Computational time cost for feasibility filtering.}
\label{tab:feasibility_filtering}
\centering
\begin{tabular}{lr}
\toprule
\textbf{Pipeline} & \textbf{Feasibility Filtering (min)} \\
\midrule
ContractEval & 1.75 \\
\bottomrule
\end{tabular}
\end{table}

\section{Sensitivity to LLM Engine Selection}\label{app:llm_engine_selection}
Various commercial models struggle with generating valid CVTs.
Table \ref{tab:avc_ts_comparison} compares AVC and TS across o4-mini, GPT-4o, and Gemini-3-flash.

The empirical results reveal a clear trade-off rather than a uniform performance advantage.
GPT-4o and Gemini-3-flash improve TS scores, achieving 78.73\% and 84.18\%, respectively, against the 74.11\% of o4-mini.
However, they concurrently experience a noticeable drop in AVC, falling from 98.68\% to 93.41\% and 95.70\%, respectively.
Because stronger models fail to maximize both metrics simultaneously, deploying them incurs unnecessarily high computational and financial costs without delivering strictly superior generation.
Therefore, we employ o4-mini as our primary LLM engine.

\begin{table}[hbt]
\caption{Comparison of AVC and TS across different models, evaluated on a 100-task sample.}
\label{tab:avc_ts_comparison}
\centering
\setlength{\tabcolsep}{8pt}
\begin{tabular}{lrr}
\toprule
\textbf{Model} & \textbf{AVC} & \textbf{TS} \\
\midrule
o4-mini & 98.68\% & 74.11\% \\
GPT-4o & 93.41\% & 78.73\% \\
Gemini-3-flash & 95.70\% & 84.18\% \\
\bottomrule
\end{tabular}
\end{table}

\section{Implementation Detail}\label{ap:implementation-detail}
We use \texttt{o4-mini} for contract-violating test generation, contract-aware query reconstruction,
and LLM-as-judge, with $\texttt{temperature}=1.0$ and $\texttt{num\_samples}=1$ per query.
For open-source code generation, we run inference with
$\texttt{batch\_size}=10$, $\texttt{num\_samples}=1$, $\texttt{max\_length}=4096$, and
$\texttt{max\_new\_tokens}=2048$.
All experiments are conducted on 2$\times$NVIDIA RTX A6000 GPUs and an AMD Ryzen Threadripper 3960X (24-core) CPU.

\section{Case Study: Logical Contradictions in Direct LLM Test Case Generation}\label{ap:qa1}
\paragraph{HumanEval} As shown in Figure~\ref{fig:tc_humaneval}, 
this task includes three sequential contracts: \texttt{assert\_0} 
checks if the input is a list, \texttt{assert\_1} 
verifies that all elements in the list are strings, 
and \texttt{assert\_2} ensures that all strings consist only of digits. 
A critical dependency exists between these contracts. 
Specifically, \texttt{assert\_2} can only be evaluated if \texttt{assert\_1} 
is satisfied, because the \texttt{isdigit()} method is only valid for string types.
A test case designed to violate \texttt{assert\_1} while satisfying 
\texttt{assert\_0} and \texttt{assert\_2} would therefore be a logically
contradictory combination, as a non-string element would cause a
\texttt{TypeError} before \texttt{assert\_2} could be checked. 
Despite this, a direct LLM generation approach often produces such invalid 
combinations. For instance, when tasked to generate test cases, the LLM produces
inputs such as \verb|[123, "456"]|, \verb|["789", [0]]|, and \verb|["456", false]|. 
These examples fail to isolate a specific contract violation.
This highlights a fundamental weakness of the approach, as the LLM tends to generate simplistic
contract-violation test cases that fail to respect the logical relationships among contracts.

\begin{figure}[hbt]
\centering
\begin{tcolorbox}[
    colback=blue!5,
    colframe=blue!40!black,
    title=\textbf{HumanEval/113},
    left=1mm, right=1mm, top=1mm, bottom=1mm,
    enhanced, sharp corners
]
\begin{lstlisting}
def odd_count(lst):
    assert type(lst) == list, "invalid inputs" # $_CONTRACT_$
    assert all(isinstance(s, str) for s in lst), "invalid inputs" # $_CONTRACT_$
    assert all(s.isdigit() for s in lst), "invalid inputs" # $_CONTRACT_$

    ans, template = [], "the number of odd elements in the string i of the input."
    for s in lst:
        odd_cnt = len(list(filter(lambda ch: int(ch) % 2 == 1, s)))
        ans.append(template.replace("i", str(odd_cnt)))
    return ans

"""
Contract List:
assert_0: assert type(lst) == list, "invalid inputs
assert_1: assert all(isinstance(s, str) for s in lst), "invalid inputs
assert_2: assert all(s.isdigit() for s in lst), "invalid inputs
""" 
\end{lstlisting}
\end{tcolorbox}
\caption{Code and contracts for HumanEval.}
\label{fig:tc_humaneval}
\end{figure}

\paragraph{MBPP} As shown in Figure~\ref{fig:tc_mbpp}, 
this task includes four main contracts, which can be grouped by their dependency.
The initial contracts, \texttt{assert\_0} and \texttt{assert\_1}, 
perform type checking to verify that both inputs are of a numeric type, such as an integer or a 
floating-point number. The subsequent contracts, \texttt{assert\_2} and \texttt{assert\_3}, 
check numeric 
properties, such as ensuring the numbers are positive or fall within a specific range.
A critical dependency exists between these groups of contracts.
Specifically, the numeric property checks in \texttt{assert\_2} and \texttt{assert\_3} can only be
evaluated if the type checks in \texttt{assert\_0} and \texttt{assert\_1} are satisfied.
For example, a non-numeric type like a \texttt{string} or \texttt{null} cannot be evaluated 
for properties like being positive. Therefore, creating a contract-violation test case that
violates the initial type contracts (\texttt{assert\_0} or \texttt{assert\_1})
while simultaneously satisfying the subsequent property contracts (\texttt{assert\_2} 
and \texttt{assert\_3}) is a logical impossibility.
Despite this, a direct LLM generation approach often produces such logically flawed combinations.
For instance, when tasked to generate test cases, the LLM produces inputs such as 
\verb|["abc", null]|, \verb|[null, "abc"]|, \verb|[[1], {"x":1}]|, 
and \verb|[{"r":1}, [2]]|. Crucially, while these examples successfully violate the initial type
contracts, they all inherently fail to satisfy \texttt{assert\_2} and \texttt{assert\_3},
yet they are generated as if such a combination were possible.
This highlights a fundamental weakness of the approach, as the LLM tends to generate 
simplistic contract-violation test cases that fail to respect the logical relationships among contracts.

\begin{figure}[hbt]
\centering
\begin{tcolorbox}[
    colback=blue!5,
    colframe=blue!40!black,
    title=\textbf{MBPP/731},
    left=1mm, right=1mm, top=1mm, bottom=1mm,
    enhanced, sharp corners
]
\begin{lstlisting}
import math
def lateralsurface_cone(r,h):

  assert isinstance(r, (int, float)), "invalid inputs" # $_CONTRACT_$
  assert isinstance(h, (int, float)), "invalid inputs" # $_CONTRACT_$
  assert r > 0, "invalid inputs" # $_CONTRACT_$
  assert h > 0, "invalid inputs" # $_CONTRACT_$
  l = math.sqrt(r * r + h * h)
  return math.pi * r  * l

"""
Contract List:
assert_0: assert isinstance(r, (int, float)), "invalid inputs
assert_1: assert isinstance(h, (int, float)), "invalid inputs
assert_2: assert r > 0, "invalid inputs
assert_3: assert h > 0, "invalid inputs
"""
\end{lstlisting}
\end{tcolorbox}
\caption{Code and contracts for MBPP.}
\label{fig:tc_mbpp}
\end{figure}

\section{Case Study: Contract Enforcement under Different Prompting Conditions}\label{ap:qa2}
We present a direct comparison of code snippets that the DeepSeek-R1 model 
generated for the \textbf{MBPP/11} task under two prompting conditions.
This comparison illustrates why providing CVTs is more effective
than relying on natural language descriptions alone.

\begin{figure}[hbt]
\centering
\begin{tcolorbox}[
    colback=blue!5,
    colframe=blue!40!black,
    title=\textbf{MBPP/11},
    left=1mm, right=1mm, top=1mm, bottom=1mm,
    enhanced, sharp corners
]
\begin{lstlisting}
"""
Write a python function that removes the first and last occurrence of a given character from the string; the first input must be a non-empty string and the second input must be a string of length one.
"""

def remove_Occ(s, char):
    assert len(s) > 0, "First input must be a non-empty string."
    assert len(char) == 1, "Second input must be a string of length one."
    
    first = s.find(char)
    last = s.rfind(char)
    
    if first == -1:
        return s
    
    return s[:first] + s[first+1:last] + s[last+1:]
\end{lstlisting}
\end{tcolorbox}
\caption{Code generated by DeepSeek with the contract specification~(CS) prompt.}
\label{fig:tc_mbpp_cs}
\end{figure}

Figure~\ref{fig:tc_mbpp_cs} shows the code snippet produced when the model received 
the contract specification~(CS) prompt that describes contracts in natural language.
The prompt contains enough information 
to infer all necessary contracts. It specifies two explicit contracts, 
namely that the first input must be a non-empty string and that the second 
input must be a string of length one. It also implies two type contracts that 
both inputs must be strings. The generated code snippet correctly implements the
explicit length-based contracts with \verb|assert len(s) > 0| and \verb|assert len(char) == 1|. However, it omits the implicit type checks and
treats the word ``string'' as descriptive context rather than a strict precondition. As a result, the function fails to enforce the contracts under non-string inputs.

In contrast, Figure~\ref{fig:tc_mbpp_ct} displays the output when the prompt
was augmented with concrete CVTs such as \verb|remove_Occ(2.0, 3.0)|.
These test cases provide an unambiguous and executable signal that non-string
inputs are invalid and must be rejected. The model generates a more robust 
implementation that now includes the crucial \verb|isinstance(s, str)| and 
\verb|isinstance(char, str)| assertions.

This comparison shows that natural language descriptions may contain all required information, yet their inherent ambiguity can prevent the model from adhering to the contract. CVTs make these implicit contracts explicit and enforceable. They compel the model to produce a code snippet that is more comprehensive and adheres to the intended contracts.

\begin{figure}[ht!]
\centering
\begin{tcolorbox}[
    colback=blue!5,
    colframe=blue!40!black,
    title=\textbf{MBPP/11},
    left=1mm, right=1mm, top=1mm, bottom=1mm,
    enhanced, sharp corners
]
\begin{lstlisting}
"""
Write a python function that removes the first and last occurrence of a given character from the string; the first input must be a non-empty string and the second input must be a string of length one.

# Contract Test Cases:
>>> remove_Occ(2.0, 3.0)
"AssertionError: invalid input"
>>> remove_Occ(5.0, 4.0)
"AssertionError: invalid input"
>>> remove_Occ('', 2.0)
"AssertionError: invalid input"
>>> remove_Occ(2.0, '')
"AssertionError: invalid input"
"""

def remove_Occ(s, char):
    assert isinstance(s, str) and len(s) > 0, "invalid input"
    assert isinstance(char, str) and len(char) == 1, "invalid input"
    
    first_occ = s.find(char)
    last_occ = s.rfind(char)
    
    if first_occ == -1:
        return s
    
    return s[:first_occ] + s[first_occ+1:last_occ] + s[last_occ+1:]
\end{lstlisting}
\end{tcolorbox}
\caption{Code generated by DeepSeek with the example-augmented specification~(EAS) prompt.}
\label{fig:tc_mbpp_ct}
\end{figure}

\section{Formalizing Contracts into SMT-LIB}\label{app:smt-template}
This section details the structure of the SMT-LIB template used in the construction of ContractEval. 
SMT-LIB is a standardized, text-based language used to interface with SMT solvers. 
It provides a formal syntax for declaring variables, defining functions, and asserting logical formulas, allowing complex problems to be translated into 
a format that a solver can systematically analyze for satisfiability.
Our framework leverages this language to translate nuanced, natural language contracts into a formal representation that can be reasoned about with logical precision.

Figure~\ref{fig:ap4_adt} shows the base template we designed for this purpose. 
It is composed of several key components, each serving a distinct role in the test generation process.
The placeholders within this template are populated as follows:
\begin{itemize}
    \item \textbf{CANONICAL PYTHON-LIKE ADT:} This fixed block defines a universal data structure for representing common Python types. 
    This allows the SMT solver to reason about various input types in a standardized way.
    \item \textbf{HELPER FUNCTIONS:} This section is populated with custom functions needed to define the contracts for a specific task.
    For example, a function to check if a string contains only digits would be defined here.
    \item \textbf{INPUT:} The input variables for the function under test are declared here.
    \item \textbf{BASIC\_STRUCTURE:} This section defines fundamental structural constraints on the inputs, such as ensuring a variable is a list composed of integer values.
    \item \textbf{CONTRACT\_DEFS:} The specific logical rules of each contract are translated into formal predicates in this section.
    \item \textbf{COMBINATION:} This is the core logic for generating a test case. 
    It contains assertions stating which contracts must be satisfied and which must be violated.
    The SMT solver then attempts to find a concrete model that satisfies this exact combination of constraints.
\end{itemize}

\begin{figure}[hbt]
\centering
\begin{tcolorbox}[
    colback=blue!5,
    colframe=blue!40!black,
    title=\textbf{The SMT-LIB template},
    left=1mm, right=1mm, top=1mm, bottom=1mm,
    enhanced, sharp corners
]

\begin{lstlisting}
ADT_BASE_TEMPLATE = """
(set-logic ALL)

; ==== CANONICAL PYTHON-LIKE ADT (DO NOT MODIFY) ====
(declare-datatypes ((Value 0)) (
  ((IntVal (ival Int))
   (FloatVal (fval Real))
   (StrVal (sval String))
   (BoolVal (bval Bool))
   (Nil)
   (Cons (head Value) (tail Value)))
))

; === ADD HELPER FUNCTIONS HERE ===
<<HELPER_FUNCTIONS>>

; === Inputs ===
<<INPUT>>

; === BASIC STRUCTURE ===
<<BASIC_STRUCTURE>>

; === Contract predicates ===
<<CONTRACT_DEFS>>

; === COMBINATION ===
<<COMBINATION>>

(check-sat)
(get-model)
"""

\end{lstlisting}
\end{tcolorbox}
\caption{The SMT-LIB template used for formalizing contracts.}
\label{fig:ap4_adt}
\end{figure}

\section{Case Study: Formalizing HumanEval/11}\label{app:smt-template-case-study}
Figure~\ref{fig:ap4_code} presents the ground-truth Python implementation for the \textbf{HumanEval/11} task. 
This specific problem requires a function accepting two binary strings of equal length. 
Figure~\ref{fig:ap4_adt_example} illustrates the corresponding SMT-LIB formalization of these requirements. 
The Python code contains three explicit assert statements, which map directly to three formal contract definitions in the SMT-LIB script:
\textbf{C0} verifies the type constraints, corresponding directly to the statement \texttt{assert isinstance(a, str) and isinstance(b, str)}.
\textbf{C1} enforces the length equality, mirroring the \texttt{assert len(a) == len(b)} condition.
\textbf{C2} evaluates the character-level validity using a custom \texttt{isBinaryString} helper function. 
This logical block perfectly translates the Python set operations (\texttt{assert set(a).issubset(\{"0", "1"\})}) 
into standard regular expression constraints (\texttt{str.in.re}) within the SMT domain.

\begin{figure}[hbt]
\centering
\begin{tcolorbox}[
    colback=blue!5,
    colframe=blue!40!black,
    title=\textbf{HumanEval/11},
    left=1mm, right=1mm, top=1mm, bottom=1mm,
    enhanced, sharp corners
]

\begin{lstlisting}
from typing import List

def string_xor(a: str, b: str) -> str:
    
    assert isinstance(a, str) and isinstance(b, str), "invalid inputs" # $_CONTRACT_$
    assert len(a) == len(b), "invalid inputs" # $_CONTRACT_$
    assert set(a).issubset({"0", "1"}) and set(b).issubset({"0", "1"}), "invalid inputs" # $_CONTRACT_$
    
    return "".join(str(int(a[i]) ^ int(b[i])) for i in range(len(a)))
\end{lstlisting}
\end{tcolorbox}
\caption{The ground-truth implementation in HumanEval/11 }
\label{fig:ap4_code}
\end{figure}

The formalization process introduces specialized helper functions to bridge the fundamental semantic gap between Python's dynamic typing and SMT-LIB's strict static typing. 
The universal ADT encapsulates various Python variables, yet interacting with these encapsulated types requires rigorous safety mechanisms. 
For instance, the \texttt{Safe\_Sval} function safely extracts string values from the custom ADT. 
By explicitly defining fallback behaviors—such as returning an empty string when encountering non-string types—this helper 
function prevents catastrophic type-matching errors during solver execution. 
Such structural safeguards guarantee that the automated pipeline remains highly resilient, even when processing complex or unexpected inputs.

The COMBINATION block dictates the ultimate objective of the test case generation by defining the exact combinatorial space of the contracts. 
By deliberately asserting either a contract predicate, such as \texttt{(assert C0)}, or its logical negation, such as \texttt{(assert (not C0))}, the pipeline commands the SMT solver to systematically explore specific constraint subsets. 
This targeted exploration isolates distinct fault domains, allowing the framework to evaluate whether a language model fails on a single edge case or across multiple interacting constraints. 
The specific instance in Figure~\ref{fig:ap4_adt_example} asserts the negation of all three contracts simultaneously. 
Consequently, the solver synthesizes a singular concrete test case that violates every precondition at once. 
This mathematically verified control allows ContractEval to generate diverse, strictly valid CVTs.


\begin{figure}[H]
\centering
\begin{tcolorbox}[
    colback=blue!5,
    colframe=blue!40!black,
    title=\textbf{The SMT-LIB template},
    left=1mm, right=1mm, top=1mm, bottom=1mm,
    enhanced, sharp corners
]

\begin{lstlisting}
ADT_BASE_TEMPLATE = """
(set-logic ALL)
; ==== CANONICAL PYTHON-LIKE ADT (DO NOT MODIFY) ====
(declare-datatypes ((Value 0)) (
  ((IntVal (ival Int))
   (FloatVal (fval Real))
   (StrVal (sval String))
   (BoolVal (bval Bool))
   (Nil)
   (Cons (head Value) (tail Value)))
))
; === ADD HELPER FUNCTIONS HERE ===
(define-fun Safe_Sval ((x Value)) String
  (ite (is-StrVal x) (sval x) ""))
(define-fun isBinaryString ((s Value)) Bool
  (and (is-StrVal s)
       (str.in.re (Safe_Sval s) (re.* (re.union (str.to.re "0") (str.to.re "1"))))))
; === Inputs ===
(declare-const a Value)
(declare-const b Value)
; === BASIC STRUCTURE ===

; === Contract predicates ===
(define-fun C0 () Bool (and (is-StrVal a) (is-StrVal b)))
(define-fun C1 () Bool (= (str.len (Safe_Sval a)) (str.len (Safe_Sval b))))
(define-fun C2 () Bool (and (isBinaryString a) (isBinaryString b)))
; === COMBINATION ===
(assert (not C0))
(assert (not C1))
(assert (not C2))

(check-sat)
(get-model)
"""
\end{lstlisting}
\end{tcolorbox}
\caption{An example of the SMT-LIB template populated for HumanEval/11}
\label{fig:ap4_adt_example}
\end{figure}

\section{Full Results of Prompting Strategies}\label{app:prompting-full}
Table~\ref{tab:full_prompting_result} reports the full metrics for the prompting variants
discussed in Section~\ref{ssec:rq-contract-ct-in-codegen}.
Beyond \texttt{pass@1} and CSR, we include CodeBLEU and LLM-as-judge to capture
syntactic and semantic alignment of the generated contract checks with the reference.
Overall, CoT prompting is high-variance and often harmful:
it can substantially degrade both functional correctness and contract satisfaction,
and it catastrophically collapses for Phi-4 and Phi-4-plus (near-zero across all metrics).
In contrast, the multi-turn strategy is markedly more stable.
It typically recovers \texttt{pass@1} relative to the single-turn EAS baseline
(e.g., up to +4.2\%p for Phi-4-plus), while keeping CSR competitive
(within roughly 0--7\%p of EAS across models).
The auxiliary metrics largely track CSR:
when CSR drops under a prompting variant, CodeBLEU and LLM-as-judge tend to drop as well,
suggesting that performance changes are driven by missing or altered contract checks
rather than superficial formatting differences.

\begin{table*}[hbt]
\caption{Comparison of additional prompting strategies.
Multi-turn recovers pass@1 while maintaining competitive contract satisfaction with the standard EAS prompt.
}
\label{tab:full_prompting_result}
\centering
\begin{tabular}{llrrrr}
\toprule
\multirow{2}{*}{\textbf{Model}} & \multirow{2}{*}{\textbf{Instruction}} & \multicolumn{1}{c}{\textbf{Functional}} & \multicolumn{3}{c}{\textbf{Contract Satisfaction}} \\
\cmidrule(lr){3-3} \cmidrule(lr){4-6}
 & & \multicolumn{1}{c}{pass@1 ($\uparrow$)} & \multicolumn{1}{c}{CSR ($\uparrow$)} & \multicolumn{1}{c}{CodeBLEU ($\uparrow$)} & \multicolumn{1}{c}{LLM-as-judge ($\uparrow$)} \\\midrule
\multirow{3}{*}{DeepSeek-R1}                  & EAS              & 70.87\%          & \textbf{52.66\%} & \textbf{54.51\%} & \textbf{81.49\%} \\
                                              & + CoT            & 61.11\%          & 35.00\%          & 40.35\%          & 62.46\%          \\
                                              & +Multi-Turn      & \textbf{72.96\%} & 45.44\%          & 45.75\%          & 65.34\%          \\\midrule
\multirow{3}{*}{gemma-3}                      & EAS              & 69.56\%          & 48.54\%          & 48.88\%          & 72.04\%          \\
                                              & + CoT            & 70.46\%          & 50.37\%          & 50.68\%          & 71.23\%          \\
                                              & +Multi-Turn      & \textbf{71.34\%} & \textbf{49.26\%} & \textbf{49.62\%} & \textbf{72.45\%} \\\midrule
\multirow{3}{*}{Phi-4}                        & EAS              & \textbf{72.99\%} & \textbf{50.73\%} & \textbf{50.86\%} & \textbf{75.63\%} \\
                                              & + CoT            & 3.66\%           & 1.91\%           & 8.14\%           & 13.28\%          \\
                                              & +Multi-Turn      & 71.88\%          & 46.58\%          & 48.44\%          & 73.42\%          \\\midrule
\multirow{3}{*}{Phi-4-plus}                   & EAS              & 73.21\%          & \textbf{50.94\%} & \textbf{51.04\%} & \textbf{75.98\%} \\
                                              & + CoT            & 0.00\%           & 0.00\%           & 2.92\%           & 4.96\%           \\
                                              & +Multi-Turn      & \textbf{77.42\%} & 48.52\%          & 49.87\%          & 75.01\%          \\\midrule
\multirow{3}{*}{Qwen-3}                       & EAS              & 69.65\%          & \textbf{51.83\%} & \textbf{53.30\%} & \textbf{77.27\%} \\
                                              & + CoT            & 71.13\%          & 45.24\%          & 50.01\%          & 75.62\%          \\
                                              & +Multi-Turn      & \textbf{71.70\%} & 49.23\%          & 50.97\%          & 76.98\%          \\\bottomrule
\end{tabular}
\end{table*}

\section{Contract Evaluation on Functionally Correct Programs}
\label{app:contract-evaluation-correct-function}

\begin{figure*}[hbt]
    \centering
    \includegraphics[width=.97\linewidth]{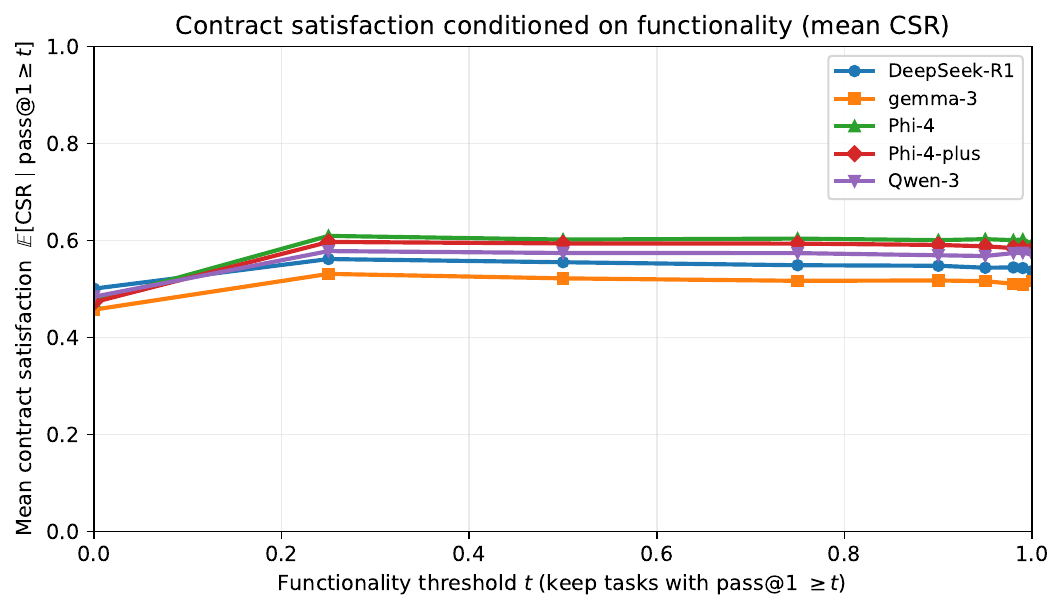}
    \caption{Contract satisfaction depending on the performance of functional correctness.}
    \label{fig:contract-condition-function}
\end{figure*}

Through ContractEval, our experiments show that prompting LLMs with contract information~(e.g., CS and EAS)
substantially improves contract satisfaction on ill-formed tests.
However, contract satisfaction is only meaningful when the generated program is also
\emph{functionally correct} on well-formed inputs.
If a program already fails on valid inputs, then its behavior on ill-formed inputs is ambiguous:
it may ``reject'' an input simply because it crashes (e.g., due to unrelated bugs), rather than
because it implements the intended contract checks.
We disentangle true contract behavior from incidental failures
by analyzing contract satisfaction \emph{conditioned on functional correctness}
in both Figure~\ref{fig:contract-condition-function} and Table~\ref{tab:contract-with-perfect-function}.

Let $f_i \in [0,1]$ denote the functional test pass rate of a generated program for task $i$
(measured on well-formed tests), and let $c_i \in [0,1]$ denote its contract satisfaction
(measured by CSR on ContractEval's ill-formed tests).
For a functionality threshold $t$, we consider the subset
$\mathcal{S}_t = \{ i \mid f_i \ge t \}$ and report:
(i) the conditional mean contract satisfaction $\mathbb{E}[c \mid f \ge t]$, and
(ii) the probability of perfect contract satisfaction $\Pr(c=1 \mid f \ge t)$.
Intuitively, increasing $t$ filters to progressively more functionally reliable programs, and
reveals whether contract satisfaction improves as functional correctness improves.

Across models, we observe that contract satisfaction does \emph{not} reliably track functional correctness.
After removing the most functionally incorrect outputs (e.g., $t \ge 0.25$),
$\mathbb{E}[c \mid f \ge t]$ becomes largely flat: making a program more functionally correct
does not automatically make it more contract-satisfying.
Even among \emph{fully} functionally correct programs ($f=1$),
a substantial fraction still fails to reject contract-violating inputs.
For example, under our contract-aware prompting setting,
only about half of functionally correct programs achieve perfect contract satisfaction
($\Pr(c=1 \mid f=1)$ is roughly in the 51\%--59\% range across models),
indicating that many solutions that ``solve the task'' still omit essential validity checks.

This conditional analysis supports the central premise of ContractEval:
functional correctness on well-formed inputs is insufficient to characterize robustness.
Contract satisfaction remains a distinct and non-trivial dimension of code quality,
and ContractEval's ill-formed tests expose failures that are invisible under pass@k-style evaluation.

\begin{table*}[hbt]
\caption{Contract satisfaction evaluation on functionally correct programs.
}
\label{tab:contract-with-perfect-function}
\centering
\begin{tabular}{llrrrr}
\toprule
\multirow{2}{*}{\textbf{Model}} & \multirow{2}{*}{\textbf{Instruction}} & \multicolumn{1}{c}{\textbf{Correct Functions}} & \multicolumn{3}{c}{\textbf{Contract Satisfaction}} \\
\cmidrule(lr){3-3} \cmidrule(lr){4-6}
 & & \multicolumn{1}{c}{Ratio ($\uparrow$)} & \multicolumn{1}{c}{CSR ($\uparrow$)} & \multicolumn{1}{c}{CodeBLEU ($\uparrow$)} & \multicolumn{1}{c}{LLM-as-judge ($\uparrow$)} \\\midrule
\multirow{2}{*}{DeepSeek-R1}                  & CS            & \textbf{63.19}\%                           & 32.89\%          & 35.96\%          & 55.38\%          \\
                                              & EAS           & 59.34\%                                    & \textbf{56.75\%} & \textbf{55.01\%} & \textbf{83.01\%} \\\midrule
\multirow{2}{*}{gemma-3}                      & CS            & \textbf{64.84}\%                           & 26.53\%          & 31.98\%          & 44.61\%          \\
                                              & EAS           & 57.14\%                                    & \textbf{54.29\%} & \textbf{54.20\%} & \textbf{79.99\%} \\\midrule
\multirow{2}{*}{Phi-4}                        & CS            & 61.26\%                                    & 52.01\%          & 52.25\%          & 78.34\%          \\
                                              & EAS           & \textbf{61.81}\%                           & \textbf{61.78\%} & \textbf{56.33\%} & \textbf{84.21\%} \\\midrule
\multirow{3}{*}{Phi-4-plus}                   & CS            & 59.89\%                                    & 49.58\%          & 48.11\%          & 71.89\%          \\
                                              & EAS           & \textbf{62.91}\%                           & \textbf{61.82\%} & \textbf{55.31\%} & \textbf{83.31\%} \\\midrule
\multirow{2}{*}{Qwen-3}                       & CS            & \textbf{60.99}\%                           & 30.05\%          & 39.93\%          & 55.83\%          \\
                                              & EAS           & 59.34\%                                    & \textbf{58.76\%} & \textbf{57.79\%} & \textbf{85.18\%} \\\bottomrule
\end{tabular}
\end{table*}

\end{document}